\documentclass[10pt,twocolumn,letterpaper]{article}

\usepackage[pagenumbers]{cvpr} 

\usepackage[dvipsnames]{xcolor}

\definecolor{cvprblue}{rgb}{0.21,0.49,0.74}
\usepackage[pagebackref,breaklinks,colorlinks,citecolor=cvprblue]{hyperref}

\usepackage{tabularx}
\usepackage{color}
\usepackage{colortbl}
\usepackage{xcolor}
\usepackage{multirow}
\usepackage{multicol}
\usepackage{wrapfig}

\def\paperID{9906} 
\def\confName{CVPR}
\def\confYear{2024}

\title{Bridging the Gap: A Unified Video Comprehension Framework \\for Moment Retrieval and Highlight Detection}

\author{Yicheng Xiao$^{1}$\thanks{Equal contribution.
  $\dagger$ Corresponding author.}, \quad Zhuoyan Luo$^{1\ast}$\\
  Yong Liu$^{1}$, Yue Ma$^{1}$, Hengwei Bian$^{2}$, Yatai Ji$^{1}$, Yujiu Yang$^{1\dagger}$, Xiu Li$^{1\dagger}$ \\
  $^{1}$Tsinghua Shenzhen International Graduate School, Tsinghua University\\
  $^{2}$Carnegie Mellon University\\
{\tt\small \{xiaoyc23, luozy23\}@mails.tsinghua.edu.cn}
}

\begin{document}
\maketitle
\begin{abstract}
Video Moment Retrieval (MR) and Highlight Detection (HD) have attracted significant attention due to the growing demand for video analysis.
Recent approaches treat MR and HD as similar video grounding problems and address them together with transformer-based architecture.
However, we observe that the emphasis of MR and HD differs, with one necessitating the perception of local relationships and the other prioritizing the understanding of global contexts.
Consequently, the lack of task-specific design will inevitably lead to limitations in associating the intrinsic specialty of two tasks.
To tackle the issue, we propose a \textbf{U}nified \textbf{V}ideo \textbf{COM}prehension framework (UVCOM) to bridge the gap and jointly solve MR and HD effectively.
By performing progressive integration on intra and inter-modality across multi-granularity, UVCOM achieves the comprehensive understanding in processing a video.
Moreover, we present multi-aspect contrastive learning to consolidate the local relation modeling and global knowledge accumulation via well aligned multi-modal space.
Extensive experiments on QVHighlights, Charades-STA, TACoS, YouTube Highlights and TVSum datasets demonstrate the effectiveness and rationality of UVCOM which outperforms the state-of-the-art methods by a remarkable margin.
Code is available at \href{https://github.com/EasonXiao-888/UVCOM}{https://github.com/EasonXiao-888/UVCOM}

\end{abstract}    
\section{Introduction}
\label{sec:intro}
Video has emerged as a highly favored multi-medium format on the internet with its diverse content.
This significant surge in online video encourages users to adjust their strategies for accessing desired video contents.
Instead of spending time-consuming efforts inspecting the whole video, they are more inclined to directly obtain particular clips of interest through language descriptions.
This shift in user preference gives rise to two significant research topics: Video Moment Retrieval~\cite{ctrl,early_mr_1,man,recent_mr_1,recent_mr_2}, focuses on locating the specific moment, and Highlight Detection~\cite{video2gif,Yao_2016,cvs,LIM-s,joint_va}, is dedicated to identifying segments of high saliency.

Actually, it is apparent that two tasks share many common characteristics, \textit{e.g.}, identifying relevant video segments in response to textual expressions.   
In light of the above, Lei \textit{et.al.}~\cite{momentdetr} first proposes a novel dataset named QVHighlights and a basic framework called Moment-DETR to jointly solve both tasks.
UMT~\cite{umt} incorporates extra audio modality and QD-DETR~\cite{qddetr} produces text-query dependent video representation to achieve better performance.
\begin{figure}
    \centering
    \includegraphics[width=\linewidth]{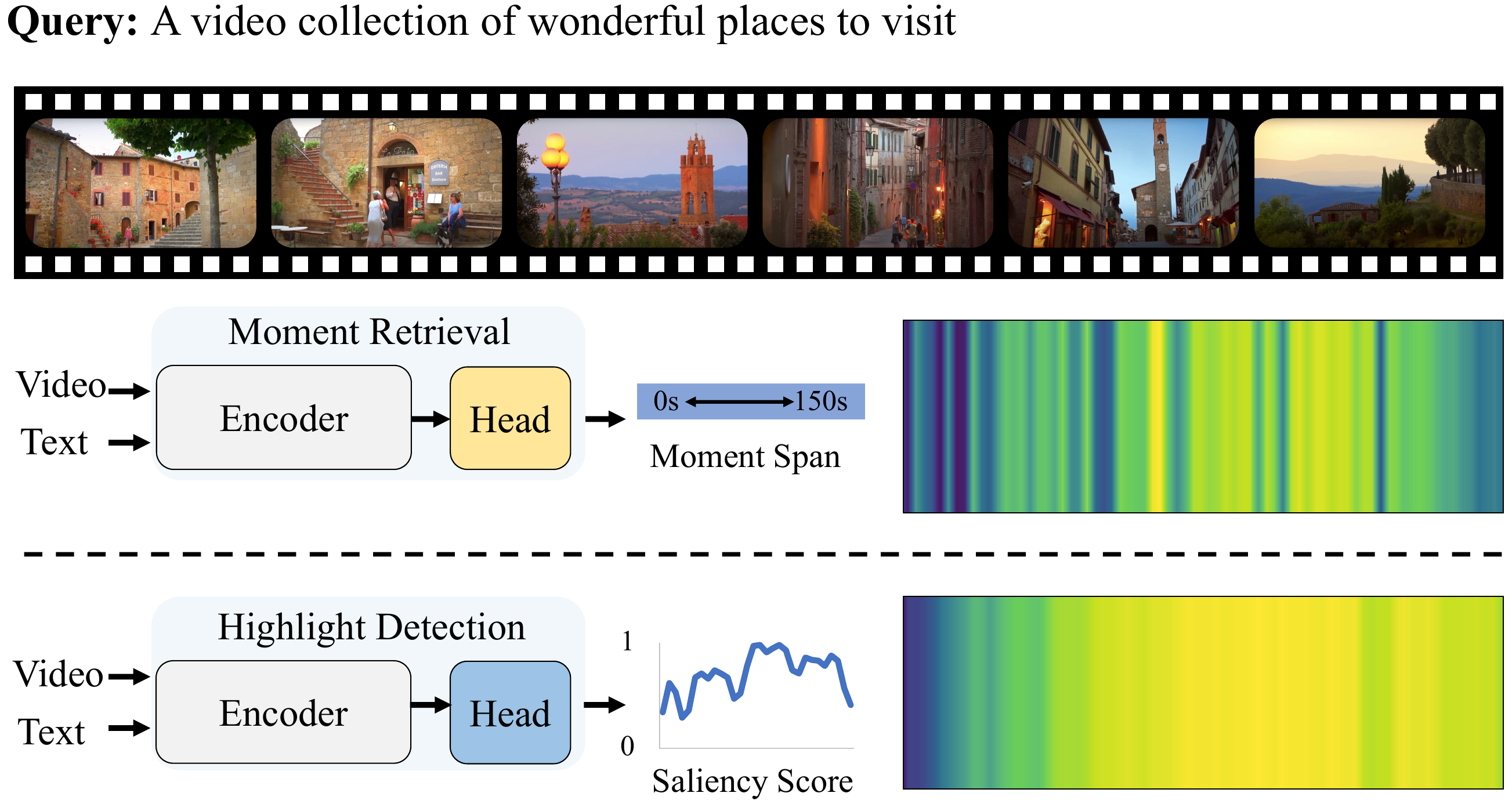}
    \caption{\textbf{Illustration of the intrinsic characteristics of Moment Retrieval and Highlight Detection.} We visualize the attention map of the same video under two tasks. 
    The attention map for MR takes on strip-like patterns, indicating the emphasis of local relations. In contrast, it is in band-like format for HD, which signifies the focus on global information.}
    \label{fig:teaser}
    \vspace{-10pt}
\end{figure}
The above-mentioned methods simply model MR and HD as a multi-task problem and mainly concentrate on utilizing non-specific strategy to solve them. 
In particular, they all adopt a straightforward way to train and optimize both tasks together with general design, \textit{e.g.}, transformer-based models.
However, we revisit the characteristics of MR/HD and discover that there exists a gap between them as illustrated in \cref{fig:teaser}. 
which leads to  
the challenge in consistent performance on both tasks, \textit{i.e.}, achieving precise moment localization and accurate highlight-ness estimation simultaneously.

Therefore, we consider that the design of framework should follow two principles to alleviate the above weakness: 1) \textit{\textbf{Local Relation Activation:}} MR necessitates the understanding of local relationships within the video to accurately localize specific segments. 2) \textit{\textbf{Global Knowledge Accumulation:}} The objective of HD is to fit the saliency distribution of the entire video, emphasizing the importance of global context (in \cref{fig:teaser}). Based on the principles, we propose a \textbf{U}nified \textbf{V}ideo \textbf{Com}prehension Framework (UVCOM) to seamlessly integrate the emphasis of MR and HD, which effectively bridges the gap and achieves great performance on both tasks consistently. 
Specifically, we first design a novel Comprehensive Integration Module (CIM) to progressively facilitate the integration on intra and inter-modality across multi-granularity.
CIM first efforts to propagate the aggregated semantic phrases from the text into the visual feature to realize local relationship perception.
Then, it accumulates global information from video by utilizing the moment-awareness feature as an intermediary.
With a comprehensive view of the entire video, CIM facilitates the understanding of particular intervals and highlight contents, which is beneficial to identify the desired moment and non-related ones. 
Furthermore, we introduce a multi-aspect contrastive learning which incorporates clip-text alignment to consolidate the local relation modeling, and video-linguistic discrimination to enhance the quality of accumulated global information.

We conduct extensive experiments on five popular MR/HD benchmarks  
to validate the effectiveness of our framework and the results show that UVCOM notably outperforms existing methods for all benchmarks.

Overall, our contributions are summarized as follows:
\begin{itemize}
    \item Based on our investigation into the emphasis of Moment Retrieval and Highlight Detection, we present two principles for framework design. Guided by them, we propose a Unified Video Comprehension Framework called UVCOM to effectively bridge the gap between two tasks.
    \item In UVCOM, a Comprehensive Integration Module (CIM) is designed to perform progressive intra and inter-modality interaction across multi-granularity, which achieves locality perception of temporal and multi-modal relationships as well as global knowledge accumulation of the entire video.
    \item Without bells and whistles, our method outperforms all existing state-of-the-art methods by a remarkable margin, \textit{e.g.}, $+5.97$\% in R1@0.7 for MR than UniVTG~\cite{univtg} on TACoS~\cite{tacos} and $+3.31$\% in HIT@1 for HD than QD-DETR~\cite{qddetr} on QVHighlights~\cite{momentdetr}.
\end{itemize}

\begin{figure*}
    \centering
    \includegraphics[width=\linewidth]{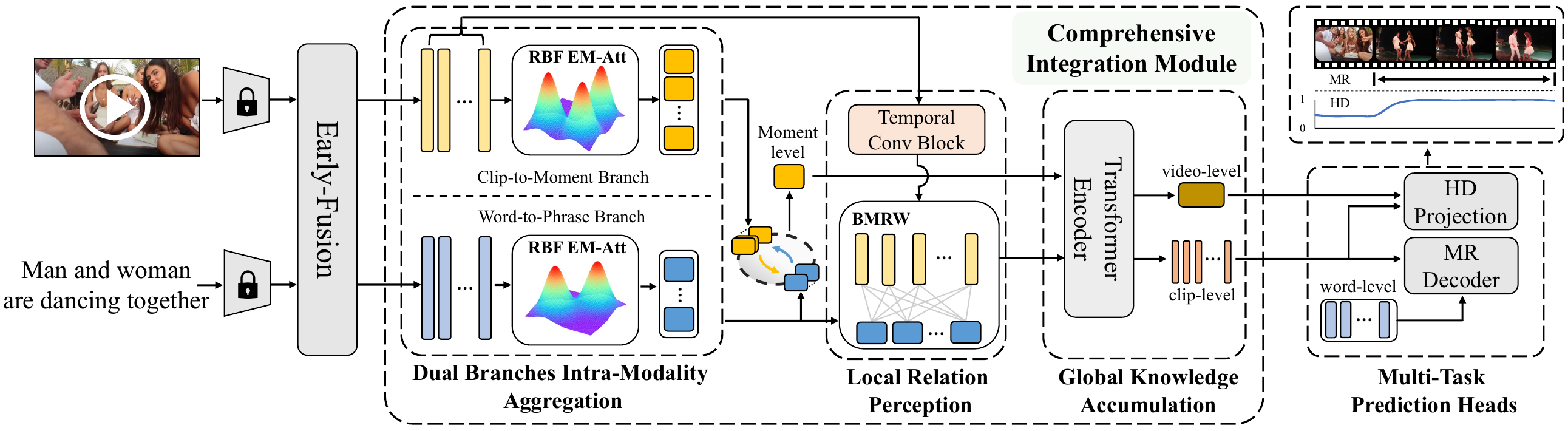}
    \caption{\textbf{Overview of UVCOM.} Based on the exploration of MR and HD, we propose a unified video comprehension framework guided by the design principles. Specifically, the model takes a video with language description as input. After encoding and early-fusion process, we design a Comprehensive Integration Module (CIM) to achieve subsequent progressive integration on intra and inter-modality across multi-granularity. Finally, the multi-task heads output the moment spans for MR and saliency scores for HD.}
    \label{fig:framework}
    \vspace{-10pt}
\end{figure*}

\section{Related Work}
\label{sec:related_work}
\paragraph{Moment Retrieval.} Moment Retrieval is a task that aims at retrieving the target moment, \textit{i.e.}, one~\cite{ctrl} or many~\cite{tvr} continuous intervals in a video given the text description. Generally, the model will focus more on the relationship across adjacent frames for better localization. Previous works retrieve video intervals into two perspectives: proposal-based and proposal-free. The proposal-based methods~\cite{hendricks, bpn, ctrl, mpn} follow the propose-then-rank pipeline, where they first generate candidate proposals then rank them based on matching scores. Liu \textit{et al.}~\cite{liumeng-2018} and Hendricks \textit{et al.}~\cite{hendricks} utilize sliding windows to scan the entire video for candidate proposals generation and calculate the similarity with textual embedding for selection. 
To enhance temporal awareness, MAN~\cite{man} and Gao \textit{et al.}~\cite{gao_2021} both introduce graph convolution network to capture temporal dependency across adjacent moments. 
On the other hand, the proposal-free methods~\cite{yuan_2019, mun_2020, rodriguez_2020, VSLNET, li_2021, Tang_2022} directly regress the start and end timestamp via video-text interaction. Yuan \textit{et al.}~\cite{yuan_2019} and Mun \textit{et al.}~\cite{mun_2020} generate the temporal coordinates of sentence by multi-modal co-attention mechanism. Furthermore, Rodriguez \textit{et al.}~\cite{rodriguez_2020} utilizes a simple dynamic filter instead of cross attention to match video and text embedding. 

\vspace{-10pt} 
\paragraph{Highlight Detection.} Highlight Detection aims to identify highlights or important segments with high potential appeal in a video. Compared with Moment Retrieval, It is necessary for the model to associate the whole video content for fitting saliency distribution of each clip. Many prior works~\cite{Yao_2016, video2gif, Yao_2016, LIM-s, cvs} adopt ranking formulation where they rank the important segments with higher score. Video2Gif~\cite{video2gif} trains a generic highlight predictor to produce GIF from videos. Rochan \textit{et al.}~\cite{Rochan_2020} designs a task-specific highlight detectors to automatically create highlights from the user history. They all use the fine-grained annotations for training which are labor-intensive and expansive. Therefore, Xiong \textit{et al.}~\cite{LIM-s} proposes to learn highlights only with video-level label. Recently, Badamdorj \textit{et al.}~\cite{joint_va} elaborates on fusing visual and audio content to generate better video representations.

MR and HD share many similar properties. Moment-DETR~\cite{momentdetr} puts forward a novel dataset which first includes two tasks and provides a simple DETR-based~\cite{detr} network. To improve the query quality, UMT~\cite{umt} proposes to adopt audio, visual and text content for query generation. Furthermore, QD-DETR~\cite{qddetr} exploits the textual information by involving video-text pair negative relationship learning, achieving greater performance. However, previous methods simply train and optimize two tasks without considering the different emphasis of each task. To address this issue, we propose a novel unified framework UVCOM that effectively associates the speciality of MR and HD to achieve comprehensive understanding. 

\section{Method}

Given a video of $L$ clips $\{v_1, v_2, \dots, v_l\}$ and a textual expression of $N$ words $\{e_1, e_2, \dots, e_n\}$, the goal of MR is to localize the most relevant moment with the center coordinate and duration, while HL is to generate the saliency score distribution for the whole video.

\subsection{Visual-Text Encoding}
\label{sec:visual-text encoding}
\paragraph{Visual Encoder.} Following previous works~\cite{momentdetr,umt,qddetr,univtg}, we utilize the pretrained backbone, \textit{e.g.}, SlowFast~\cite{slowfast}, video encoder of CLIP~\cite{clip} and I3D~\cite{i3d} to extract visual features $\mathcal{F}_v \in \mathbb{R}^{L \times D}$ of the video. Note that D demotes the channel.
\vspace{-12pt}
\paragraph{Language Encoder.} Simultaneously, text encoder of CLIP is adopted to encode the linguistic expression into the textual embedding $\mathcal{F}_{t} \in \mathbb{R}^{N \times D}$.
 
With the visual and textual features, we apply a bidirectional transformer-based encoder to perform the early fusion.
It coarsely encodes features in different modalities and outputs preliminary aligned visual and textual representations.  

\subsection{Comprehensive Integration Module}
\label{sec:cim}
After getting the visual and textual representations, we design a Comprehensive Integration Module (CIM) to perform progressive intra and inter-modality integration across multi-granularity. 
Specifically, we leverage Expectation-Maximum (EM) Attention~\cite{em} on associating inner-modality content to generate the moment-wise visual features and phrase-wise textual features, respectively. 
Then we propose Local Relation Perception (LRP) module to unify temporal relationship modeling and inter-modality fusion, which reformulates the temporal and modality inter-connection to enhance the locality perception.
Finally, we utilize a standard encoder to produce the video-wise feature by integrating the correlation between moment and clip-wise visual features.
\vspace{-10pt}
\paragraph{Dual Branches Intra-Modality Aggregation.} A video usually contains more than one event and irrelevant background scenes.
The same scenario happens in textual descriptions where insignificant words and unconstrained expressions may cause potential ambiguity. 
To tackle the problem, we propose to utilize RBF-kernel based EM Attention~\cite{em,rskp} to aggregate the clip/word-level features. 
As shown in \cref{fig:framework}, it is a dual-branches structure.
The clip-to-moment branch aims at incorporating the relationship of each clip to enhance the desired event representations while suppressing the background noise.
Meanwhile, the word-to-phrase branch is to emphasize the referred moment description by accumulating contextual information.

Specifically, we fit the distribution of $\mathcal{F}_v$ and $\mathcal{F}_t$ by a separated Gaussian Mixture Model~\cite{gmm} to generate the compact moment and phrase-level representations via the centroid of Gaussians.
Taking $\mathcal{F}_v$ as an example,
we utilize a linear superposition of $n_v$ Gaussians to capture the statistics of $f_v^i \in \mathbb{R}^D$ (the $i$-th snippet of $\mathcal{F}_v$):
\begin{equation}
p(f_v^i)=\sum_{k=1}^{n_v} z_{k}^v\mathcal{N} (f_v^i|\mu_k, \Sigma _k),
\end{equation}
where $z_{k}^v \in \mathbb{R}$, $\mu_k \in \mathbb{R}^D$ and $\Sigma_k \in \mathbb{R}^{D\times D}$ denote the weight, mean and covariance of $k$-th Gaussian basis for the clip-to-moment branch.
We substitute the covariance with an identity matrix $I$ for simplification and employ the radial basis function (RBF Kernel) $\mathcal{K}(f_v^i,\mu_k)$ to estimate the posterior probability $\mathcal{N} (f_v^i|\mu _k, \mathcal{I} )$:
\begin{equation}
    \mathcal{K}(f_v^i,\mu_k) = exp(- \lambda \left \| f_v^i -\mu_k \right \| _2^2),
\end{equation}
where $\lambda > 0$ is an adjustable hyper-parameter to control the distribution.
Afterwards, at $t$-th iteration, we update the weight $Z^{(t)} \in \mathbb{R}^{L\times n_v}$ in the E Step and re-estimate $\mu^{(t)} \in \mathbb{R}^{n_v\times D}$ in the M step, which can be formulated as:
\begin{equation}
\label{t-step}
    \mu^{(t)} = \mathrm{Norm_1}(Z^{(t)})^TF_v,\quad t\in \left \{  1,\dots,T\right \}.
\end{equation}
Furthermore, in contrast to conventional cluster methods that only involve iterative update, the initialized means $\mu^{(0)}$ we set are learnable.
Therefore, they can effectively capture the feature distribution of the dataset through the standard back-propagation.

After $t$ iterations, we obtain the fine-grained moment-wise representation $\mathcal{F}_m$ from $\mu^{(t)}$ which fully aggregates the contextual information.
Similarly, we operate the above steps on word-to-phrase branch to generate the phrase-level linguistic feature $\mathcal{F}_p \in \mathbb{R}^{n_t \times D}$, where $n_t$ indicates the number of Gaussian basis in word-to-phrase branch. 

\begin{figure}
    \centering
    \includegraphics[width=0.9\linewidth]{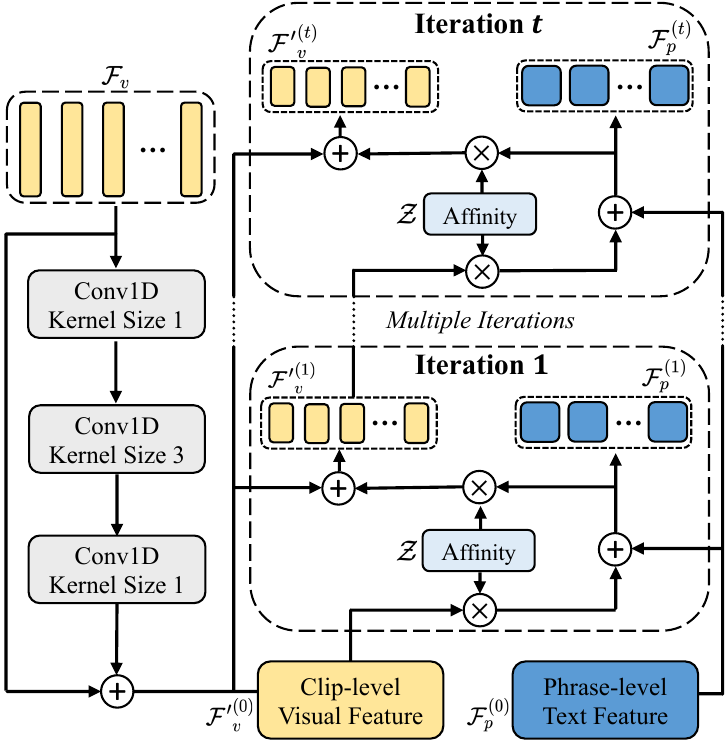}
    \caption{\textbf{Illustration of Local Relation Perception (LRP) module.} We first process the visual feature $\mathcal{F}_v$ with a Conv1D Block. Then we develop a Bidirectional Modality Random Walk (BMRW) algorithm to exploit the power of fine-grained multi-modal interaction. The affinity $\mathcal{Z}$ is generates by scaled dot product: $\mathcal{Z}=\lambda_{z}\mathcal{F'}_v^{(0)} (\mathcal{F}_p^{(0)})^{\top }$.}
    \label{fig:lrp_module}
    \vspace{-10pt}
\end{figure}
\vspace{-10pt}
\paragraph{Local Relation Perception.}
Previous methods~\cite{momentdetr, umt, qddetr} directly perform cross-modal fusion between clip and word-level features, disregarding the temporal relation and valuable semantic interaction across different granularities. Without the information from adjacent clips, the simple and coarse clip-word fusion will easily deviate the model from focusing on the relevant boundary clips, causing incorrect localization. 
To address the aforementioned weakness, we design a Local Relation Perception (LRP) module to excavate both temporal and inter-modality relationships.
As shown in \cref{fig:framework}, we first utilize a temporal convolution block to improve the locality perception of clip-level features, which can be formulated as:
\begin{equation}
    \mathcal{F'}_{v} = \mathrm{Conv}\left(\mathcal{F}_v\right) + \mathcal{F}_v.
\end{equation}
Since simply incorporate clip-level relation may introduce local redundancy, we leverage fine-grained inter-modal interaction to re-calibrate the attention for activating the relevant moments. 
Intuitively, a straightforward approach is to utilize cross-attentive mechanism~\cite{qddetr, umt} to perform inter-modal interaction.
Nevertheless, the complex scenario in an untrimmed video, \textit{e.g.}, footage transitions and irrelevant events, will increase the likelihood of attention drift which leads to the undesirable local activation.
Moreover, although phrase-wise linguistic features specify referred moment description and alleviate the impact of noise in contrast to word-wise one, it may potentially contribute to attention drift due to the irrelevant accumulated words.
Therefore, inspired by~\cite{rskp,Giannis2020,Grady2006}, we design a  
bidirectional modality random walk (BMRW) algorithm to mitigate the mentioned drawbacks and fully exploit the power of the fine-grained multi-modal interaction.
It propagates the textual prior into the visual features for highlighting the corresponding local context and suppressing unrelated ones.
Simultaneously, linguistic features are refined through the incorporation of updated visual content. 
As shown in \cref{fig:lrp_module}, there are multiple iterations in BMRW where two modalities features learn collaboratively in visual-linguistic shared embedding space until convergence.

Formally, we first define the $\mathcal{F'}_v$, $\mathcal{F}_p$ as initial features $\mathcal{F'}_v^{(0)}$, $\mathcal{F}_p^{(0)}$ at $0$-th iteration and formulate affinity $\mathcal{Z}$ by scaled dot product: $\mathcal{Z}=\lambda_{z}\mathcal{F'}_v^{(0)} (\mathcal{F}_p^{(0)})^{\top }$, where $\lambda_{z}$ is the scaling factor.
At $t$-th iteration, the phrase-wise linguistic feature $\mathcal{F}_p^{(t)}$ is updated by the original feature $\mathcal{F}_p^{(0)}$ and the visual output $\mathcal{F'}_v^{(t-1)}$ from previous iteration: 
\begin{equation}
    \label{txt_t_iter}
    \mathcal{F}_p^{(t)}=\omega \mathrm{Norm1}(\mathcal{Z})^{\top }\mathcal{F'}_v^{(t-1)}+(1-\omega )\mathcal{F}_p^{(0)},
\end{equation}
Subsequently, it is projected into the temporal-awareness feature $\mathcal{F'}_{v}^{(t)}$:
\begin{equation}
    \label{vid_t_iter}
    \mathcal{F'}_v^{(t)}=\omega \mathcal{Z}\mathcal{F}_p^{(t)} + (1-\omega )\mathcal{F'}_v^{(0)},
\end{equation}
where $\omega \in (0,1)$ is the factor which controls the degree of modalities fusion.
Then, we substitute $\mathcal{F}_p^{(t)}$ into \cref{vid_t_iter} to derive the iterative update formula of $\mathcal{F'}_v^{(t)}$:
\begin{equation}
    \mathcal{F'}_v^{(t)} = (\omega^2A)^{t} \mathcal{F'}_v^{(0)} + (1-\omega )\sum_{i=0}^{t-1} (\omega^2A)^{i}(\omega \mathcal{Z}\mathcal{F}_p^{(0)}+\mathcal{F'}_v^{(0)}),
\end{equation}
where $A$ denotes $\mathcal{Z}\mathrm{Norm1}(\mathcal{Z})^{\top}$. Intuitively, the moment-specific regions of visual features can be fully activated by the guidance of textual features after multiple iterations.
Moreover, to avoid the potential issue of unexpected gradient and high computation cost, we use an approximate inference function based on Neumann Series~\cite{neuman} when $t \to \infty$:
\begin{equation}
    \label{new_vid}
    \mathcal{F'}_v^{(\infty )}=(1-\omega )(I-\omega ^2 A)^{-1}(\omega \mathcal{Z} \mathcal{F}_p^{(0)}+ \mathcal{F'}_v^{(0)}).
\end{equation}

In this manner, the model realizes a synergistic temporal and inter-modality relation integration and generates a more comprehensive visual representation $\mathcal{F}_v^{new}$, \textit{i.e.}, $\mathcal{F'}_v^{(\infty )}$ in~\cref{new_vid}.

\vspace{-10pt}
\paragraph{Global Knowledge Accumulation.}
As illustrated in \cref{fig:teaser}, Highlight Detection prioritizes global information of videos.
QD-DETR~\cite{qddetr} uses a saliency token to capture general information. 
However, the input-agonist design might cause the inferior perception of the text-related intervals due to the non-referential search area.
To mitigate the concern, we propose to use the moment-aware feature as intermediate guidance to accumulate the global knowledge of a video.  
Specifically, we derive the most relevant snippet $\mathcal{F'}_m$ by measuring the similarity between the moment-wise $\mathcal{F}_m $ and phrase-wise embeddings $\mathcal{F}_p$. Then, a stack of transformer encoder layers~\cite{transformer} are utilized to excavate the correlation between $\mathcal{F'}_m$ and $\mathcal{F}_v^{new}$. The overall process is:
\begin{equation}
    \mathcal{F}_v^{g}, \mathcal{F}_v^{l} = Encoder(Concat[ \mathcal{F'}_m,\mathcal{F}_v^{new}]).
\end{equation}
Consequently, the semantic snippet is obliged to focus on the referred moment and suppress the non-target response, which eventually produces the video-wise feature $\mathcal{F}_v^{g} \in \mathbb{R}^{1\times D}$. In addition, $\mathcal{F}_v^{l} \in \mathbb{R}^{L\times D}$ is greatly enriched by the supplement of global information.  

\subsection{Multi-Aspect Contrastive Learning}
\label{sec:MACL}
As discussed in \cref{sec:cim}, CIM can better accomplish local relation enhancement in temporal and inter-modality as well as global knowledge accumulation of a video. It is anticipated that the explicit supervision of each objective will further consolidate the effectiveness. To this end, we introduce multi-aspect contrastive learning, which is in two folds:  
\paragraph{Clip-Text Alignment.} This loss bridges the semantic gap between the textual expression and the clip-level features, which further improves the quality of local relation modeling. Specifically, we first average $\mathcal{F}_t$ to get the sentence-level textual embedding $\mathcal{F'}_t \in \mathbb{R}^{1 \times D}$ and then measure the relevance with clip-level visual representation $\mathcal{F}_v^{new}$:
\begin{equation}
    S_{ct} = \frac{\mathcal{F}_v^{new}\cdot \mathcal{F'}_t^{\top } }{\left \| \mathcal{F}_v^{new} \right \|\cdot \left \|  \mathcal{F'}_t \right \| }. 
\label{cta}
\end{equation}
Finally, we compute the contrastive loss by matrix multiplication:
\begin{equation}
    \mathcal{L}_{cta} = -\mathrm{LogSoftmax}\left(S_{ct}\right)\cdot G_{ct},
\label{cta_loss}
\end{equation}
where $G_{ct}$ is annotated to $1$ for relevant clips and $0$ for others.
\paragraph{Video-Linguist Discrimination.} It aims at constructing the fine-grained multi-modal joint space where video-level visual feature $\{\mathcal{F}_{v(i)}^g\}_{i=1}^{B}$ closens relevant sentence-level textual representation $\{\mathcal{F'}_{t(i)}\}_{i=1}^{B}$ while distances unrelated ones within a batch $B$.
Similar to~\cite{clip}, the whole process can be formulated as:  
\begin{equation}
    \mathcal{L}_{vld} = -\sum_{i=1}^{B}\mathrm{Log}\frac{exp\left(\mathcal{F}_{v(i)}^g \cdot \mathcal{F'}_{t(i)}^{\top}\right)}{\sum_{j=1}^B exp\left(\mathcal{F}_{v(j)}^g \cdot \mathcal{F'}_{t(j)}^{\top}\right)}.
\end{equation}

\subsection{Prediction Heads and Loss Function}
\label{sec:phlf}
\paragraph{Multi-Task Prediction Heads.}
As depicted in~\cref{fig:framework}, there are two simple heads built on top of the Comprehensive Integration Module for Moment Retrieval and Highlight Detection respectively.
Similar to~\cite{momentdetr,qddetr,eatr}, Moment Retrieval Head comprises a standard transformer decoder where we leverage $\mathcal{F}_t$ as the query to generate a series of moment spans $P_m$.
Highlight Detection Head consists of two groups of single fully-connected layer for linear projection.
Accordingly, we get the prediction saliency scores $P_{s} \in \mathbb{R}^{L\times 1} $:
\begin{equation}
    P_s = \frac{\mathcal{F}_v^{g} w_g^{\top } \cdot\mathcal{F}_v^{l} w_l^{\top }}{\sqrt[]{d} },
\end{equation}
where $w_{g}$ and $w_{l} \in \mathbb{R}^{d\times D}$ are learnable weights of two Layers.

\paragraph{Total Loss.}
We supervise our framework by four groups of training objective functions.
For MR, $L1$ loss and $GIoU$ loss are adopted to measure the disparity between GT moment $G_m$ and prediction spans $P_m$ : 
\begin{equation}
    \mathcal{L}_{MR} = \lambda_{gIoU}\mathcal{L}_{gIoU}(P_m,G_m) + \lambda_{L1}\mathcal{L}_{L1}(P_m,G_m).
\end{equation}
Moreover, the loss functions for HD consist of margin ranking loss $\mathcal{L}_{margin}$ and rank-aware loss $\mathcal{L}_{rank}$ following~\cite{qddetr}. Both losses work in tandem to ensure the predicted saliency scores $P_s$ conform to the ground truth scores $G_s$ :
\begin{equation}
    \mathcal{L}_{HD} = \lambda_{HD}\left[\mathcal{L}_{margin}(P_s,G_s) + \mathcal{L}_{rank}(P_s,G_s)]\right.
\end{equation}
Inspired by~\cite{qddetr, dong2017}, we involve hard samples into training process for diversifying the formulations of local and global relationships of different video-text pairs. Briefly, we categorize the lowest relevance between video and text as hard samples and suppress their saliency scores $P_s^{hard}$ during training:   
\begin{equation}
    \mathcal{L}_{hard} = -\lambda_{hard}\mathrm{Log}(1-P_s^{hard})
\end{equation}

\noindent In addition, the objective of multi-aspect contrastive learning promotes semantic associations between text descriptions and visual contents of multi-granularity:
\begin{equation}
    \mathcal{L}_{con} = \lambda_{cta}\mathcal{L}_{cta} + \lambda_{vld}\mathcal{L}_{vld}.
\end{equation}
Generally, the total loss is expressed as:
\begin{equation}
    \mathcal{L}_{total} = \mathcal{L}_{HD} + \mathcal{L}_{MR} + \mathcal{L}_{hard} + \mathcal{L}_{con}.
\end{equation}
The $\lambda$ above are hyper-parameters for balancing the losses.
\section{Experiments}

\subsection{Datasets and Evaluation Metrics}
\paragraph{Datasets.} We evaluate our model on five prevalent MR/HD benchmarks: QVHighlights~\cite{momentdetr}, Charades-STA~\cite{cha}, TaCoS~\cite{tacos}, TVSum~\cite{tvsum} and YouTube Highlights~\cite{youtubehl}. Due to the space limitation, the details of each datasets are included in the~\cref{sup:datasets}.

\vspace{-7pt}
\paragraph{Metrics.} 
Following~\cite{momentdetr,umt,detr}, we measure the performance of our model by the same criteria for QVhighlights, Charades-STA, TACoS, YouTube Highlights and TVSum. For descriptions of the metrics corresponding to datasets, please see the~\cref{sup:metrics}.

\subsection{Implementation Details}
\paragraph{Pre-extracted Features.} For a fair comparison, we take the same features of video, text and audio from corresponding pretrained feature extractors, \textit{e.g.}, SlowFast~\cite{slowfast}, CLIP~\cite{clip}, PANN~\cite{pann}. For more details please refer to~\cref{sup:features}.

\begin{table}[t]
\footnotesize
\setlength{\tabcolsep}{0pt}
\begin{tabularx}{\linewidth}{@{\hspace{0.1cm}}p{2.2cm}p{0.85cm}<{\centering}p{0.85cm}<{\centering}p{1.0mm}<{\centering}p{0.85cm}<{\centering}p{0.85cm}<{\centering}p{0.85cm}<{\centering}p{1.0mm}<{\centering}p{0.85cm}<{\centering}p{0.8cm}<{\centering}}
\toprule
& \multicolumn{6}{c}{\textbf{MR}} & & \multicolumn{2}{c}{\textbf{HD}} \\
\cmidrule{2-7} \cmidrule{9-10}
& \multicolumn{2}{c}{R$1$} & & \multicolumn{3}{c}{mAP} & & \multicolumn{2}{c}{$\geq$ Very Good} \\
\cmidrule{2-3} \cmidrule{5-7} \cmidrule{9-10}
\vspace{-0.73cm}\hspace{0.1cm}\textbf{Method} & @$0.5$ & @$0.7$ & & @$0.5$ & @$0.75$ & Avg. & & mAP & HIT@$1$ \\
\midrule
M-DETR \cite{momentdetr} & $52.89$ & $33.02$ & & $54.82$ & $29.40$ & $30.73$ & & $35.69$ & $55.60$ \\
{UMT}$\dagger$~\cite{umt} & $56.23$ & $41.18$ & & $53.83$ & $37.01$ & $36.12$ & & $38.18$ & $59.99$ \\
{UniVTG}~\cite{univtg} & $58.86$ & $40.86$ & & $57.60$ & $35.59$ & $35.47$ & & $38.20$ & $60.96$ \\
{MH-DETR}~\cite{mhdetr} & $60.05$ & $42.28$ & & $60.75$ & $38.13$ & $38.38$ & & $38.22$ & $60.51$ \\
{QD-DETR$\dagger$}~\cite{qddetr} & $63.06$ & $45.10$ & & $63.04$ & $40.1$ & $40.19$ & & $39.04$ & $62.87$ \\
{EaTR}~\cite{eatr} & $61.36$ & $45.79$ & & $61.86$ &  $41.91$ & $41.74$ & & $37.15$ & $58.65$ \\
\rowcolor{gray!10}
{UVCOM} & $\mathbf{63.55}$ & $\mathbf{47.47}$ & & $\mathbf{63.37}$ & $\mathbf{42.67}$ & $\mathbf{43.18}$ & & $\mathbf{39.74}$ & $\mathbf{64.20}$ \\
\rowcolor{gray!10}
{UVCOM $\dagger$} & $\mathbf{63.81}$ & $\mathbf{48.70}$ & & $\mathbf{64.47}$ & $\mathbf{44.01}$ & $\mathbf{43.27}$ & & $\mathbf{39.79}$ & $\mathbf{64.79}$ \\
\midrule
\multicolumn{8}{c}{\hspace{2cm}\textit{With ASR Captions Pretrain}} \\
\midrule
M-DETR \cite{momentdetr} & $59.78$ & $40.33$ & & $60.51$ & $35.36$ & $36.14$ & & $37.43$ & $60.17$ \\
{UMT}~\cite{umt} & $60.83$ & $43.26$ & & $57.33$ & $39.12$ & $38.08$ & & $39.12$ & $62.39$ \\
{QD-DETR}~\cite{qddetr} & $64.10$ & $46.10$ & & $64.30$ & $40.50$ & $40.62$ & & $38.52$ & $62.27$ \\
\rowcolor{gray!10}
{UVCOM } & $\mathbf{64.53}$ & $\mathbf{48.31}$ & & $\mathbf{64.78}$ & $\mathbf{43.65}$ & $\mathbf{43.80}$ & & $\mathbf{39.98}$ & $\mathbf{65.58}$ \\
\bottomrule
\end{tabularx}
\captionsetup{font={small}}

\caption{\textbf{Jointly MR and HD results on QVHighlights test split.} $\dagger$ indicates training with audio modality. \textit{With ASR Caption Pretrain} denotes models pretrained on ASR captions~\cite{momentdetr}.}

\vspace{-2pt}
\label{tab:qvhl}
\end{table}
\begin{table}[t]
\footnotesize
     \setlength{\tabcolsep}{0pt}
    \begin{tabularx}{\linewidth}{@{\hspace{0.1cm}}p{2.0cm}|@{\hspace{0.1cm}}p{1.47cm}
<{\centering}p{1.47cm}<{\centering}p{0.1cm}|@{\hspace{0.1cm}}p{1.47cm}
<{\centering}p{1.47cm}
<{\centering}p{1.47cm}
<{\centering}}
    \toprule
    \multirow{2}{*}{\vspace{-0.1cm}\textbf{Method}}  & \multicolumn{2}{c}{\textbf{Charades-STA}} &  & \multicolumn{2}{c}{\textbf{TACoS}}
    \\
     \cmidrule{2-3} \cmidrule(l{0.2cm}){4-6}
      & R1@0.5 & R1@0.7  & & R1@0.5 & R1@0.7 \\
    \midrule
    2D TAN~\cite{2d-tan}  & $46.02$ & $27.50$  & & $27.99$ & $12.92$ \\
    VSLNet~\cite{VSLNET} & $42.69$ & $24.14$  & & $23.54$ & $13.15$ \\
    M-DETR~\cite{momentdetr} & $53.63$ & $31.37$ & & $24.67$ & $11.97$ \\
    QD-DETR~\cite{qddetr} & $57.31$ & $32.55$  & & -- & -- \\
    UniVTG~\cite{univtg} & $58.01$ & $35.65$ & & $34.97$ & $17.35$ \\
    \midrule
    \rowcolor{gray!10}
    UVCOM & $\mathbf{59.25}$ & $\mathbf{36.64}$ & & $\mathbf{36.39}$ & $\mathbf{23.32}$ \\
    \bottomrule
    \end{tabularx}
    \makeatother\caption{\textbf{MR results on Charades-STA test split and TACoS test split}. The pre-extracted features are from SlowFast~\cite{slowfast} and CLIP~\cite{clip}.}
    \label{tab:tacos_cha}
\vspace{-4mm}
\end{table}
\begin{figure*}[t]
 \begin{minipage}[t]{0.4\textwidth}
  \centering
  \scriptsize
\renewcommand\tabcolsep{0pt}
\footnotesize
\begin{tabularx}{\linewidth}{@{\hspace{1mm}}p{2cm}|@{\hspace{0.5mm}}p{0.7cm}<{\centering}p{0.7cm}<{\centering}p{0.7cm}<{\centering}p{0.7cm}<{\centering}p{0.7cm}<{\centering}p{0.7cm}<{\centering}p{0.7cm}<{\centering}}
\toprule
\textbf{Method} & {Dog} & {Gym.} & {Par.} & {Ska.} & {Ski.} & {Sur.} & \textbf{Avg.} \\
\midrule
GIFs \cite{video2gif} & $30.8$ & $33.5$ & $54.0$ & $55.4$ & $32.8$ & $54.1$ & $46.4$ \\
LSVM \cite{youtubehl} & $60.0$ & $41.0$ & $61.0$ & $62.0$ & $36.0$ & $61.0$ & $53.6$ \\
LIM-S \cite{LIM-s} & $57.9$ & $41.7$ & $67.0$ & $57.8$ & $48.6$ & $65.1$ & $56.4$ \\
SL-Module \cite{sl_module} & ${70.8}$ & ${53.2}$ & $77.2$ & ${72.5}$ & ${66.1}$ &${76.2}$ & ${69.3}$ \\
MINI-Net$\dagger$~\cite{mn} & $58.2$ & $61.7$ & $70.2$ & $72.2$ & $58.7$ & $65.1$ & $64.4$ \\
TCG$\dagger$ \cite{TCG} & $55.4$ & $62.7$ & $70.9$ & $69.1$ & $60.1$ & $59.8$ & $63.0$ \\
Joint-VA$\dagger$ \cite{joint_va} & $64.5$ & $71.9$ & $80.8$ & $62.0$ & $73.2$ & $78.3$ & $71.8$ \\
{UMT}$\dagger$\cite{umt} & $65.9$ & $75.2$ & ${81.6}$ & $71.8$ & $72.3$ & ${82.7}$ & $74.9$ \\ 
{UniVTG}~\cite{univtg} &  ${71.8}$ & ${76.5}$ & $73.9$ & ${73.3}$ & ${73.2}$ & ${82.2}$ & ${75.2}$ \\ 
\midrule
\rowcolor{gray!10}
{UVCOM$^{1}$} &  $\mathbf{73.8} $ & $\mathbf{77.1}$ & ${75.7}$ & $\mathbf{75.3}$ & $\mathbf{74.0}$ & $\mathbf{82.7}$ & $\mathbf{76.4}$  \\
\rowcolor{gray!10}
{UVCOM$^{2}$} &  $\mathbf{66.5} $ & $\mathbf{77.4}$ & $\mathbf{82.8}$ & $\mathbf{78.7}$ & $\mathbf{74.2}$ & $\mathbf{84.6}$ & $\mathbf{77.4}$  \\
\bottomrule
\end{tabularx}
\vspace{-1em}
\captionsetup{font={small}}
\makeatletter\def\@captype{table}\makeatother\caption{\small{\textbf{HD results of mAP on YouTube HL.} $\dagger$ denotes using audio modality.} $1$ and $2$ indicate using the same visual and textual features of UniVTG and UMT.}
\label{tab:youtube}
  \end{minipage}
  \hspace{0.65cm}
  \begin{minipage}[t]{0.558\textwidth}
   \centering
   \scriptsize
\renewcommand\tabcolsep{0pt}
\footnotesize
\begin{tabularx}{1\linewidth}{@{\hspace{1mm}}p{1.9cm}|@{\hspace{0.4mm}}p{0.7cm}<{\centering}p{0.7cm}<{\centering}p{0.7cm}<{\centering}p{0.7cm}<{\centering}p{0.7cm}<{\centering}p{0.7cm}<{\centering}p{0.7cm}<{\centering}p{0.7cm}<{\centering}p{0.7cm}<{\centering}p{0.7cm}<{\centering}p{0.7cm}<{\centering}}
\toprule
\textbf{Method} &  {VT} & {VU} & {GA} & {MS} & {PK} & {PR} & {FM} & {BK} & {BT} & {DS} & \textbf{Avg.} \\
\midrule
sLSTM \cite{sLSTM} & ${41.1}$ & ${46.2}$ & $46.3$ & $47.7$ & $44.8$ & $46.1$ & $45.2$ & $40.6$ & $47.1$ & $45.5$ & $45.1$ \\
LIM-S \cite{LIM-s} & $55.9$ & $42.9$ & $61.2$ & $54.0$ & $60.4$ & $47.5$ & $43.2$ & $66.3$ & $69.1$ & $62.6$ & $56.3$ \\
Trailer \cite{trailer} &$61.3$ &$54.6$ & $65.7$ & $60.8$ & $59.1$ & ${70.1}$ & $58.2$ & $64.7$ & $65.6$ & ${68.1}$ & $62.8$ \\
SL-Module \cite{sl_module} & ${86.5}$ & ${68.7}$ & ${74.9}$ & $\underline{86.2}$ & ${79.0}$ & $63.2$ & ${58.9}$ & ${72.6}$ & ${78.9}$ & $64.0$ & ${73.3}$ \\
MINI-Net$\dagger$ \cite{mn} & $80.6$ & $68.3$ & $78.2$ & $81.8$ & $78.1$ & $65.8$ & $57.8$ & $75.0$ & $80.2$ & $65.5$ & $73.2$ \\
TCG$\dagger$ \cite{TCG} & $85.0$ & $71.4$ & $81.9$ & $78.6$ & $80.2$ & $75.5$ & $71.6$ & $77.3$ & $78.6$ & $68.1$ & $76.8$ \\
Joint-VA$\dagger$ \cite{joint_va} & $83.7$ & $57.3$ & $78.5$ & ${86.1}$ & $80.1$ & $69.2$ & $70.0$ & $73.0$ & $\mathbf{97.4}$ & $67.5$ & $76.3$ \\
{UniVTG}~\cite{univtg}& $83.9$ & $85.1$ & $\underline{89.0}$ & ${80.1}$ & ${84.6}$ & ${81.4}$ & ${70.9}$ & $\underline{91.7}$ & ${73.5}$ & ${69.3}$ & ${81.0}$\\
{UMT}$\dagger$\cite{umt} & ${87.5}$ & ${81.5}$ & ${88.2}$ & $78.8$ & ${81.5}$ & $\mathbf{87.0}$ & ${76.0}$ & ${86.9}$ & ${84.4}$ & $\mathbf{79.6}$ &
${83.1}$ \\
QD-DETR \cite{qddetr} & $\mathbf{88.2}$ & $\underline{87.4}$ & $85.6$ & $85.0$ & $\underline{85.8}$ & $86.9$ & $\underline{76.4}$ & $91.3$ & $\underline{89.2}$ & $73.7$ & $\underline{85.0}$ \\
\midrule
\rowcolor{gray!10}
{UVCOM} &  $\underline{87.6}$ & $\mathbf{91.6}$ & $\mathbf{91.4}$ & $\mathbf{86.7}$ & $\mathbf{86.9}$ & $\underline{86.9}$ & $\mathbf{76.9}$ & $\mathbf{92.3}$ & ${87.4}$ & $\underline{75.6}$ & $\mathbf{86.3}$\\
\bottomrule
\end{tabularx}
\vspace{-1em}
\captionsetup{font={small}}
\makeatletter\def\@captype{table}\makeatother\caption{\small{\textbf{HD results of Top-5 mAP on TVSum.} $\dagger$ denotes using audio modality. The $2$-nd performance values are highlighted by \underline{underline}.}}
\label{tab:tvsum}
\end{minipage}
\vspace{-1em}
\end{figure*}
\begin{table*}[t]
\vspace{-12pt}
\begin{minipage}[c]{\textwidth}
    \begin{minipage}{0.44\textwidth}
    \makeatletter\def\@captype{table}
    \centering
    \footnotesize
    \setlength{\tabcolsep}{4.2pt}
    \begin{tabular}{c c c@{\hspace{0.4cm}} c c c c c c}
    \toprule
    \multirow{3}{*}{\vspace{-0.2cm}\textbf{CIM}} & &\multirow{3}{*}{\vspace{-0.2cm}\textbf{MCL}} & \multicolumn{3}{c}{\textbf{MR}} & & \multicolumn{2}{c}{\textbf{HD}}
    \\
     \cmidrule{4-6} \cmidrule{8-9}
    & & & R1 & R1 & mAP & & \multirow{2}{*}{mAP} & \multirow{2}{*}{HIT@1} \\
    & & & @0.5 & @0.7 & Avg. & & & \\
    \midrule
    & & & $61.55$ & $44.84$ & $40.08$ & & $37.10$ & $62.0$ \\
    \checkmark & & & $62.84$ & $48.77$ & $43.6$ & & $39.33$ & $62.97$ \\
     & & \checkmark  & $60.77$ & $44.06$ & $40.48$ & & $38.81$ & $62.06$ \\
    \checkmark & & \checkmark & $\mathbf{65.10}$ & $\mathbf{51.81}$ & $\mathbf{45.79}$ & & $\mathbf{40.03}$ & $\mathbf{63.29}$ \\
    \bottomrule
    \end{tabular}
    \caption{\textbf{Effectiveness of the proposed modules.}}
    \label{tab:ablation_modules}
    \end{minipage}
    \begin{minipage}{0.53\textwidth}
    \vspace{10pt}
    \makeatletter\def\@captype{table}
    \centering
    \footnotesize
    \setlength{\tabcolsep}{4.3pt}
    \hspace{-2mm}
    \begin{tabular}{c c c c c c c c c}
    \toprule
    \multirow{3}{*}{\vspace{-0.2cm}\textbf{DBIA}} & \multirow{3}{*}{\vspace{-0.2cm}\textbf{LRP}} & \multirow{3}{*}{\vspace{-0.2cm}\textbf{GKA}}& \multicolumn{3}{c}{\textbf{MR}} & & \multicolumn{2}{c}{\textbf{HD}}
    \\
     \cmidrule{4-6} \cmidrule{8-9}
    & & & R1 & R1 & mAP & & \multirow{2}{*}{mAP} & \multirow{2}{*}{HIT@1} \\
    & & & @0.5 & @0.7 & Avg. & & & \\
    \midrule
    &  &  & $60.77$ & $44.06$ & $40.48$ & & $38.81$ & $62.06$ \\
     &  &\checkmark & $62.32$ & $46.71$ & $41.03$ & & $38.73$ & $62.58$ \\
      & \checkmark & & $62.06$ & $46.45$ & $41.42$ & & $38.57$ & $62.45$ \\
    \checkmark & & \checkmark & $63.74$ & $49.16$ & $43.45$ & & $39.54$ & $\mathbf{64.26}$ \\
    \checkmark& \checkmark &  & $64.71$ & $50.0$ & $43.69$ & & $39.69$ & $63.16$ \\
     & \checkmark &\checkmark & $64.84$ & $50.0$ & $44.02$ & & $39.58$ & $64.13$ \\
    \checkmark& \checkmark & \checkmark & $\mathbf{65.10} $ & $\mathbf{51.81} $ & $\mathbf{45.79} $ & & $\mathbf{40.03}$ & $63.29 $ \\
    \bottomrule
    \end{tabular}
    \hspace{-0.05\textwidth}
    \caption{\textbf{Effects of the components designed of proposed CIM module.}}
    \label{tab:ablation_CIM}
    \end{minipage}
    \begin{minipage}{0.45\textwidth}
    \makeatletter\def\@captype{table}
    \centering
    \footnotesize
    \setlength{\tabcolsep}{4.2pt}
    \vspace{-2.5mm}
    \begin{tabular}{c@{\hspace{0.4cm}} c c@{\hspace{0.5cm}} c c c c c c}
    \toprule
    & \multirow{3}{*}{\vspace{-0.2cm}\textbf{Method}} & & \multicolumn{3}{c}{\textbf{MR}} & & \multicolumn{2}{c}{\textbf{HD}}
    \\
     \cmidrule{4-6} \cmidrule{8-9}
      & & & R1 & R1 & mAP & & \multirow{2}{*}{mAP} & \multirow{2}{*}{HIT@1} \\
      & & & @0.5 & @0.7 & Avg. & & & \\
    \midrule
    & Average & & $63.48$ & $49.87$ & $44.10$ & & $39.81$ & $63.16$ \\
     & K-Means & & $62.58$ & $48.39$ & $43.13$ & & $39.47$ & $62.26$ \\
    & EM-Att & & $64.32$ & $50.26$ & $44.49$ & & $39.82$ & $\mathbf{64.0}$ \\
    & EM-Att$\dagger$ & & $\mathbf{65.10}$ & $\mathbf{51.81} $ & $\mathbf{45.79}$ & & $\mathbf{40.03}$ & $63.29$ \\
    \bottomrule
    \end{tabular}
    \caption{\textbf{Impact of various aggregation methods.} $\dagger$ indicates the EM Attention module with RBF kernel. }
    \vspace{5pt}
    \label{tab:cluster}
    \end{minipage}
    \hspace{0.025\textwidth}
    \begin{minipage}{0.51\textwidth}
    \makeatletter\def\@captype{table}
    \centering
    \footnotesize
    \vspace{-1.5mm}
    \setlength{\tabcolsep}{4.2pt}
    \begin{tabular}{c@{\hspace{0.4cm}} c c@{\hspace{0.4cm}} c c c c c c}
    \toprule
    & \multirow{3}{*}{\vspace{-0.2cm}\textbf{Method}} & & \multicolumn{3}{c}{\textbf{MR}} & & \multicolumn{2}{c}{\textbf{HD}}
    \\
     \cmidrule{4-6} \cmidrule{8-9}
      & & & R1 & R1 & mAP & & \multirow{2}{*}{mAP} & \multirow{2}{*}{HIT@1} \\
      & & & @0.5 & @0.7 & Avg. & & & \\
    \midrule
    & Cross Attention & & $63.03$ & $49.87$ & $43.79$ & & $39.63$ & $\mathbf{63.94}$ \\
    & BMRW & & $\mathbf{65.10} $ & $\mathbf{51.81} $ & $\mathbf{45.79} $ & & $\mathbf{40.03}$ & $63.29 $ \\
    \bottomrule
    \end{tabular}
    \caption{\textbf{Comparison of different modality interaction strategies.}}
    \label{tab:ablation_lrp}
    \end{minipage}
\end{minipage}
\vspace{-20pt}
\end{table*}

\vspace{-7pt}
\paragraph{Training Settings.} Our model is trained with AdamW optimizer where the learning rate is $1 \times 10^{-4}$ and weight decay is $1 \times 10^{-4}$ by default. The encoder of Global Knowledge Accumulation and the decoder of Moment Retrieval Head compose of three layers of transformer blocks.
The coefficients for losses are set to 
$\lambda_{cta}=0.5,\lambda_{hard}=1,\lambda_{vld}=0.5,\lambda_{HD}=1,\lambda_{gIoU}=1,\lambda_{L1}=10$ in default. 
Due to space limitations, please see~\cref{sup:training details} for more training details.

\subsection{Main Result}
\paragraph{QVHighlights.} We compare our method to previous methods on QVHighlights in \cref{tab:qvhl}.
Benefiting from the comprehensive understanding of the video, our UVCOM achieves new state-of-the-art performance on different settings and shows a significant margin across all metrics.
Specifically, our approach outperforms EaTR~\cite{eatr} by $2.25$\% on the average of all metrics.
Incorporating with video and audio modality, UVCOM yields a clear improvement of $3.6\%$ in R1@0.7, $4\%$ in mAP@0.75 for MR and $2\%$ in HID@1 for HD compared to QD-DETR~\cite{qddetr}.
Furthermore, with ASR caption pretraining, UVCOM achieves the greatest performance on more stringent metrics, \textit{e.g.}, $43.8\%$ in Avg. mAP for MR and $39.98$\% in Avg. mAP for HD, demonstrating the effectiveness of our method. 

\vspace{-10pt}
\paragraph{Charades-STA \& TACoS.}
In order to evaluate the performance of our method in precise moment localization, we report the results on Charades-STA and TACoS benchmarks.
As depicted in \cref{tab:tacos_cha}, UVCOM outperforms QD-DETR~\cite{qddetr} by about $4\%$ R1@0.7 using SlowFast and CLIP features in Charades-STA dataset while boosts $6\%$ R1@0.7 than UniVTG~\cite{univtg} in TaCoS. It is worth noting that we also validate our model surpasses the existing SOTA methods using VGG features (see in~\cref{sup:result_cha_vgg}).

\vspace{-10pt}
\paragraph{YouTube Highlights \& TVSum.}
For Video Highlight Detection, we conduct experiments on TVSum and YouTube Highlights. Considering the fact that the scale and scoring criteria of TVSum is small and inconsistent, our method gains incoherently among domains.
However, in~\cref{tab:tvsum}, it still boost an improvement of $1.3\%$ in Avg. mAP compared with the SOTA methods. As shown in \cref{tab:youtube}, our method achieves $76.4$\% and $77.4$\% in Avg. mAP without audio source under different settings. 
Note that the features used in UniVTG~\cite{univtg} and UMT~\cite{umt} on YouTube Highlights are different. Therefore, we follow the same protocol of each for a fair comparison.

\begin{figure*}[h]
\vspace{3mm}
    \centering
    \includegraphics[width=\linewidth]{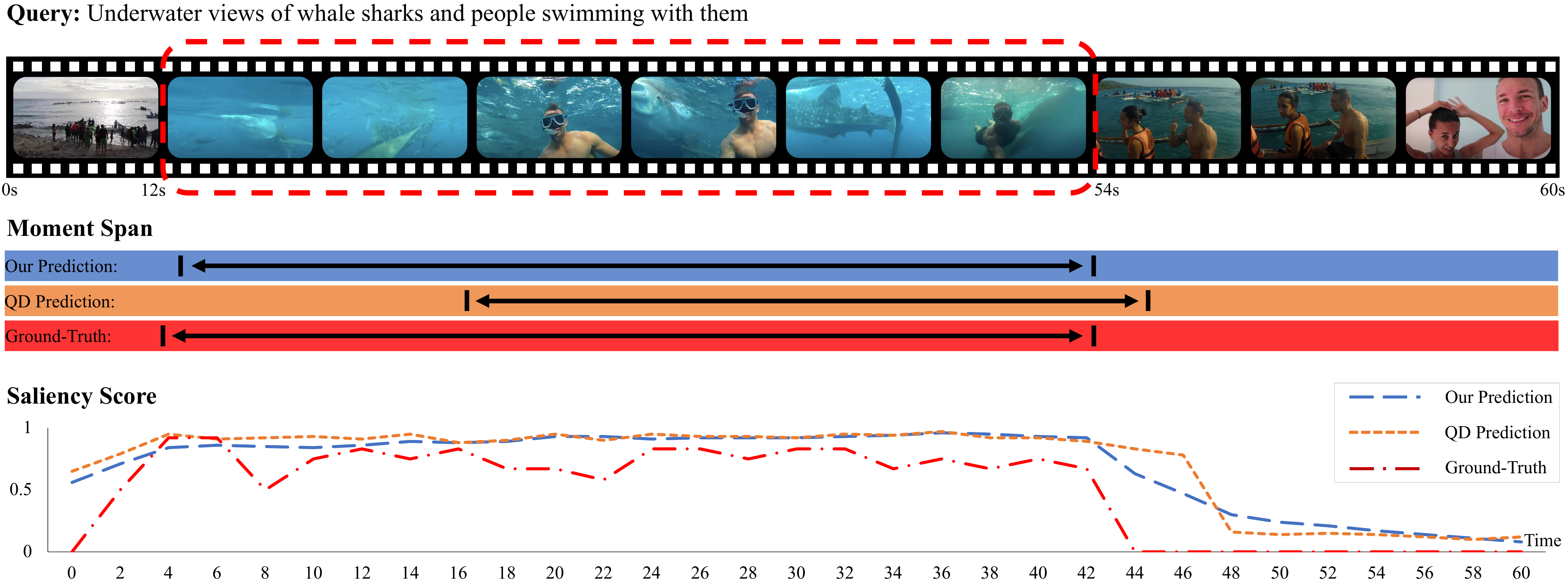}
    \caption{\textbf{Visullization comparison on MR and HD.} QD indicates previous state-of-the-art method QD-DETR~\cite{qddetr}}
    \label{fig:case_show}
    \vspace{-10pt}
\end{figure*}
\begin{figure}
    \centering
    \includegraphics[width=\linewidth]{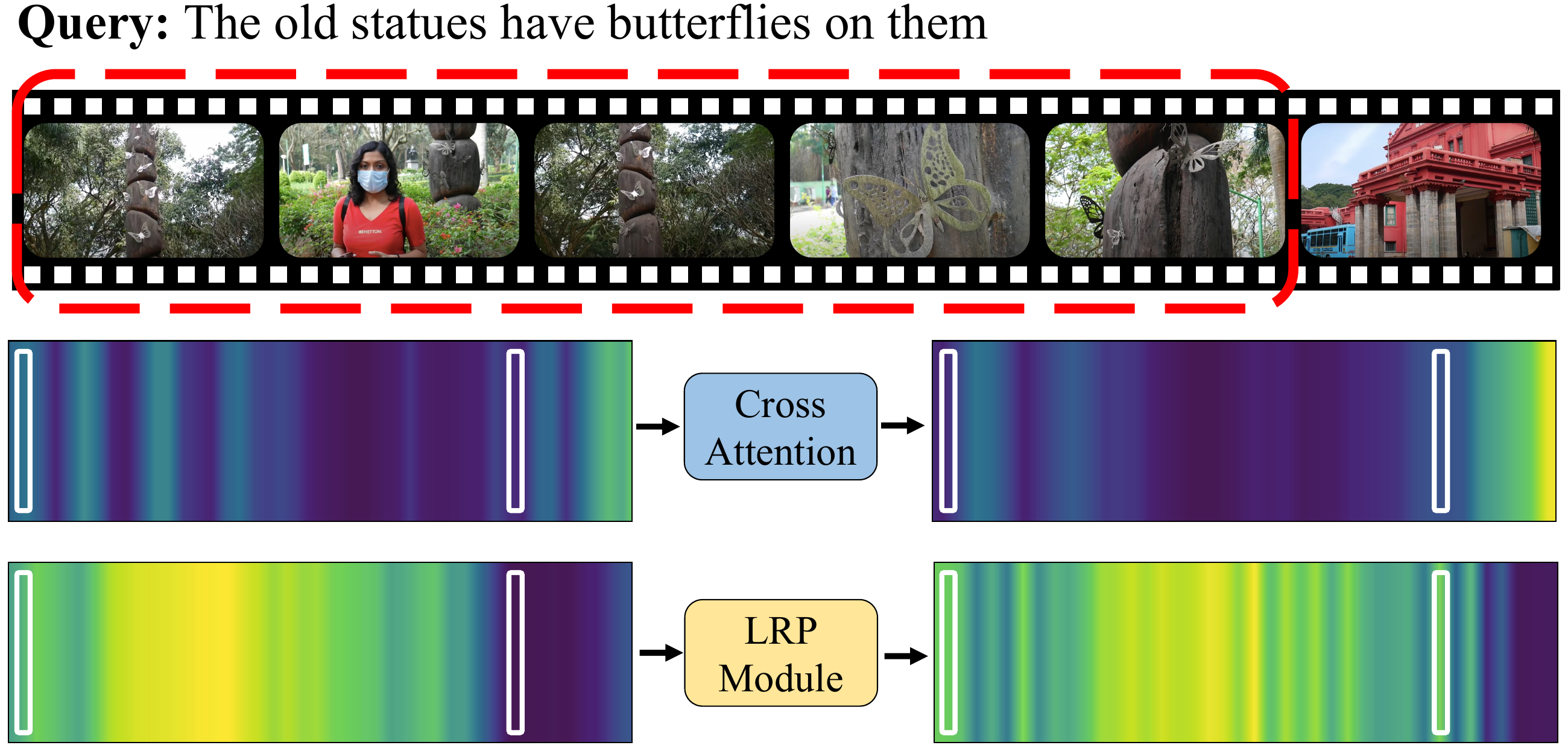}
    \caption{\textbf{Illustration of different modality interaction strategies.}
    The red bounding box indicates the relevant interval and the white bounding box denotes the start and end clips.}
    \label{fig:ablation-lrp}
    \vspace{-15pt}
\end{figure}

\subsection{Ablation Study}
In this section, we conduct a series of analysis experiments on the val split of QVHighlights benchmark and train the model from scratch without audio modality.

\paragraph{Component Analysis.}
We first verify the effectiveness of the proposed Comprehensive Integration Module (CIM) and Multi-Aspect Contrastive Learning (MCL).
As illustrated in \cref{tab:ablation_modules}, both of them brings improvement and their combination contributes to better performance ,\textit{i.e.}, $+5.71$\% in Avg. mAP, which demonstrates the effectiveness of the comprehensive understanding.
To further investigate the validity of three modules involved in CIM, we provide additional experiments on Dual Branches Intra-Modality Aggregation (DBIA), Local Relation Perception (LRP) and Global Knowledge Accumulation (GKA).  
As shown in \cref{tab:ablation_CIM}, since GKA facilitate the understanding of global context, the ablation of it leads to inferior performance on HD, \textit{i.e.}, $-1.1\%$ in HIT@1. Moreover, LRP brings a clear improvement of $+2.34\%$ in Avg. mAP on MR, proving the enhancement on locality perception.  

\vspace{-12pt}
\paragraph{Aggregation Method.}
We study the impacts on various aggregation methods utilized in DBIA module.
As illustrated in \cref{tab:cluster}, we believe the superiority of our RBF kernel based EM-Attention derives from two aspects: 1) Compared with ``Average" and K-Means, our method
enhances the desired moment representation while suppresses noises.
2) RBF kernel maps features into a high-dimensional latent space while modeling the relationship within it, which is beneficial for the subsequent aggregation.

\vspace{-12pt}
\paragraph{Modality Interaction Strategy.}
We investigate the effects of different modality interaction strategies in Local Relation Perception. As shown in \cref{tab:ablation_lrp}, replacing BMRW by cross attention mechanism results in $2\%$ performance degradation, which demonstrates the effectiveness of BMRW.
Furthermore, we provide visualization of features to prove the rationality of LRP. It can be seen in \cref{fig:ablation-lrp} that the utilization of cross-attentive mechanism leads to the emergence of attention drift. In contrast, through iterative multi-modal learning in shared space, BMRW mitigates the issue, thereby facilitating more precise localization.
Moreover, LRP achieves the local relation perception evidenced by clearer strip-like attention patterns in~\cref{fig:ablation-lrp}.

\begin{figure}[t]
    \centering
    \includegraphics[width=\linewidth]{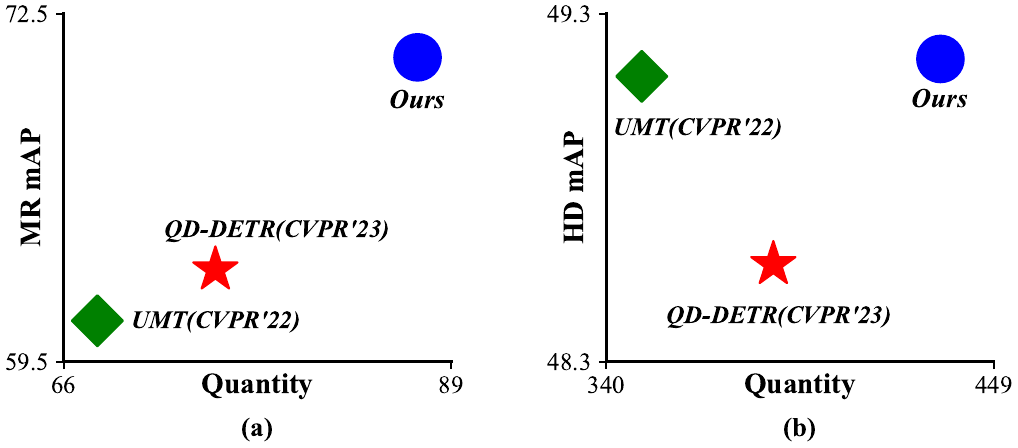}
    \caption{\textbf{Illustration of grounding consistency of MR and HD.} (a) indicates the videos collected by $mAP_{\text{HD}} > 0.8$. (b) indicates the videos collected by $mAP_{\text{MR}} > 0.8$.}
    \label{fig:gap}
    \vspace{-10pt}
\end{figure}

\paragraph{Grounding Consistency.}
Benefiting from the task-specific design, our method yields greater consistency in the joint solution of MR and HD.
To quantify the performance coherence, on one hand, we count the videos with accurate hightlight-ness estimation ($mAP_{\text{HD}} > 0.8$) and calculate MR mAP for those videos as shown in~\cref{fig:gap} (a). 
On the other hand, we measure the HD mAP and quantities of videos with precise moment spans ($mAP_{\text{MR}} > 0.8$) as shown in \cref{fig:gap} (b). 
The results demonstrate that UVCOM effectively bridges the gap between two tasks for which our method is superior on all statistics, \textit{i.e.}, MR and HD precision as well as quantity.

\subsection{Qualitative Results}
As shown in \cref{fig:case_show}, The local-global enhancement and comprehensive understanding allows our method to accurately model the saliency distribution and localize timestamps of the moment precisely. Comparatively, without the explicit association of characteristics of two tasks, QD-DETR~\cite{qddetr} struggles to handle simultaneously in complex scenarios.  

\section{Conclusion}
In light of the different emphasis on MR and HD, we propose a unified video comprehension framework called UVCOM under the guidance of design principles to effectively bridge the gap between two tasks.
By performing progressive intra and inter-modality interaction across multi-granularity, UVCOM achieves locality perception of temporal and multi-modal relationship as well as global knowledge accumulation of the entire video.
Moreover, we introduce multi-aspect contrastive learning to provide the explicit supervision of above two objectives. 
Extensive studies validate our model’s comprehensive understanding of videos and show our UVCOM remarkably outperforms the existing state-of-the-art methods.

\noindent \textbf{Limitations.} Our UVCOM has achieved comprehensive understanding of the untrimmed video and superior performance.
However, there are some potential limitations, \textit{e.g.}, unable to process auditory descriptions well.
Since we just use a simple way to handle audio features instead of specific design, we think that the explicit design for audio features is an interesting future direction.

{
    \small
    \bibliographystyle{ieeenat_fullname}
    \bibliography{main}
}
\clearpage
\setcounter{page}{1}
\appendix
\section*{\large Appendix}
\section{Additional Details about Bidirectional Modality Random Walk Algorithm.}
As stated in the manuscript, we propose a Bidirectional Modality Random Walk algorithm (BMRW) to exploit the power of the fine-grained multi-modal interaction. In this section, we provide in-depth illustration on it. First, we define the linguistic feature $\mathcal{F}_p^{(0)}$ and visual feature $\mathcal{F'}_v^{(0)}$ as initial variables at $0$-th iteration:
\begin{equation}
    \left\{\begin{matrix}
\mathcal{F}_p^{(0)}=\mathcal{F}_p
 \\
\mathcal{F'}_v^{(0)}=\mathcal{F'}_v
 \\
\mathcal{Z}=\lambda_{z}\mathcal{F'}_v^{(0)} (\mathcal{F}_p^{(0)})^{\top },
\end{matrix}\right.
\end{equation}
where $\mathcal{Z}$ is the affinity between two modalities.
Then we propagate the semantics between modalities in two iterative formulas:
\begin{equation}
    \left\{\begin{matrix}
\mathcal{F}_p^{(t)}=\omega \mathrm{Norm1}(\mathcal{Z})^{\top }\mathcal{F'}_v^{(t-1)}+(1-\omega )\mathcal{F}_p^{(0)}
 \\
\mathcal{F'}_v^{(t)}=\omega \mathcal{Z}\mathcal{F}_p^{(t)} + (1-\omega )\mathcal{F'}_v^{(0)}, \quad \omega \in (0,1).
\end{matrix}\right.
\label{equa:iteration}
\end{equation}
Integrating~\cref{equa:iteration} we can get:
\begin{equation}
\begin{aligned}
    \mathcal{F'}_v^{(t)} &= (\omega^2 ZNorm1(Z)^\top) \mathcal{F'}_v^{(t-1)} \\& \quad+ \omega(1-\omega) \mathcal{Z}\mathcal{F}_p^{(0)} + (1-\omega)\mathcal{F'}_v^{(0)} 
\\ &= (\omega^2 ZNorm1(Z)^\top) \mathcal{F'}_v^{(t-1)} \\& \quad+(1-\omega)(\omega Z \mathcal{F}_p^{(0)} + \mathcal{F'}_v^{(0)}).
\end{aligned}
\label{equa:intermediate}
\end{equation}
We substitute $ZNorm1(Z)^\top$ as $A$ and expand~\cref{equa:intermediate} from $t$-th to $0$-th iteration:
\begin{equation}
\begin{split}
    \mathcal{F'}_v^{(t)} = (\omega^2A)^{t} \mathcal{F'}_v^{(0)} + (1-\omega )\sum_{i=0}^{t-1} (\omega^2A)^{i}(\omega \mathcal{Z}\mathcal{F}_p^{(0)}+\mathcal{F'}_v^{(0)}),
\end{split}
\label{equa:result}
\end{equation}
Considering the potential risk of unexpected gradient or expensive computation cost, we use an approximate optimal solution based on Neumann Series~\cite{neuman}:
\begin{equation}
    \lim_{t \to \infty} \sum_{i=0}^{t-1} (\omega^2A)^{i} = (I -\omega ^2A)^{-1}.
\end{equation}
Accordingly, when $t \to \infty$ in \cref{equa:result}, the comprehensive visual representation is generated as:
\begin{equation}
    \mathcal{F'}_v^{(\infty )}=(1-\omega )(I-\omega ^2 A)^{-1}(\omega \mathcal{Z} \mathcal{F}_p^{(0)}+ \mathcal{F'}_v^{(0)}).
\end{equation}

\begin{table}[t]
\centering
\footnotesize
\setlength{\tabcolsep}{0pt}
\begin{tabularx}{0.7\linewidth}
{@{\hspace{0.1cm}}p{2.2cm}|@{\hspace{0.01cm}}p{1.5cm}<{\centering}p{2cm}<{\centering}p{0.2cm}}
\toprule
\textbf{Method} & {R1@0.5} & {R1@0.7} \\
\midrule
SAP~\cite{sap}  & $27.42$ & $13.36$  \\
SM-RL~\cite{sm-rl}   & $24.36$  & $11.17$ \\
MAN~\cite{man}   & $41.24$  & $20.54$ \\
2D-TAN~\cite{2d-tan}   & $40.94$  & $22.85$ \\
QD-DETR~\cite{qddetr} & $52.77$ & $31.13$ \\
\rowcolor{gray!10}
UVCOM & $ \mathbf{54.57}$ & $\mathbf{34.13}$ \\
\midrule
UMT$\dagger$~\cite{umt}   & $48.31$  & $29.25$ \\
QD-DETR$\dagger$~\cite{qddetr} & $55.51$ & $34.17$ \\
\rowcolor{gray!10}
UVCOM$\dagger$ & $ \mathbf{56.69}$ & $\mathbf{34.76}$ \\
\bottomrule
\end{tabularx}
\caption{\textbf{MR results on Charades-STA Test Split}. The pre-extracted features are from VGG, GLOVE Embeddings and PANN. $\dagger$ denotes using audio modality}
\label{tab:cha_vgg}
\end{table}
\begin{table*}[t]
\footnotesize
\centering
\vspace{10pt}
\setlength{\tabcolsep}{8.25pt}
\renewcommand{\arraystretch}{1.2}
\hspace{-1mm}
\begin{tabular}{l l c c c c c c c c c c c c}
     \toprule
     Dataset & Feature & Lr & Epoch & Bs & Lr drop & $n_v$ & $n_t$ &  $\lambda_{gIoU}$ & $\lambda_{L1}$ & $\lambda_{HD}$ & $\lambda_{hard}$ & $\lambda_{cta}$ & $\lambda_{vld}$\\
     \midrule
     QVHighlights & SF+C & 1e-4 & 200 & 32 & 100 & 30 &5 & 1 &10 & 1 &1 &0.5 &0.5 \\
     Charades-STA & SF+C & 1e-4 & 200 & 8 & 80 & 30 & 5 & 1 & 10 & 1 & 1 &0.5 &0.5 \\
     Charades-STA & VGG & 1e-4 & 200 & 8 & - & 30 & 5 & 1 & 10 & 1 & 1 &1.5 &0.5 \\
     TaCoS & SF+C & 1e-4 & 200 & 32 & 100 & 30 & 5 & 1 & 10 & 1 & 1 &0.5 &0.5 \\
     TVSum & I3D & 1e-4 & 2000 & 4 & - & 30 & 5 & 1 & 10 & 1 & 1 &\cref{tab:sup_tvsum_loss} &\cref{tab:sup_tvsum_loss} \\
     YouTubeHL & SF+C & 1e-4 & 2000 & 4 & 1000 & 30 & 2 & - & - & 1 & 1 &\cref{tab:sup_youtubehl_sfc_loss} &\cref{tab:sup_youtubehl_sfc_loss} \\
     YouTubeHL & I3D & 1e-4 & 2000 & 4 & 1000 & 30 & 2 & - & - & 1 & 1  &\cref{tab:sup_youtubehl_i3d_loss} &\cref{tab:sup_youtubehl_i3d_loss} \\
    \bottomrule
\end{tabular}
\caption{\textbf{Training details.} We provide elaborate training details on each dataset. Lr denotes learning rate; Bs denotes batch size; Lr drop denotes the drop of learning rate at the specific epoch. $n_v$ and $n_t$ denote the number of Guassians in DBIA module.}
\label{tab:sup_training_details}
\end{table*}
\begin{table*}[t]
\begin{minipage}[c]{\textwidth}
    \begin{minipage}{0.4\textwidth}
    \makeatletter\def\@captype{table}
    \centering
    \footnotesize
    \setlength{\tabcolsep}{5.5pt}
    \begin{tabular}{c c c@{\hspace{0.4cm}} c c c c c c}
    \toprule
    \multirow{3}{*}{\vspace{-0.2cm}$n_v$} & &\multirow{3}{*}{\vspace{-0.2cm}$n_t$} & \multicolumn{3}{c}{\textbf{MR}} & & \multicolumn{2}{c}{\textbf{HD}}
    \\
     \cmidrule{4-6} \cmidrule{8-9}
    & & & R1 & R1 & mAP & & \multirow{2}{*}{mAP} & \multirow{2}{*}{HIT@1} \\
    & & & @0.5 & @0.7 & Avg. & & & \\
    \midrule
    $20$ & & $3$  & $62.58$ & $48.45$ & $42.44$ & & $39.58$ & $61.87$ \\
    $25$ & & $4$  & $64.45$ & $50.06$ & $42.95$ & & $40.0$ & $\mathbf{64.9}$ \\
     $25$ & & $5$  & $64.26$ & $50.06$ & $43.48$ & & $39.51$ & $62.06$ \\
    $\mathbf{30}$ & & $\mathbf{5}$ & $\mathbf{65.10}$ & $\mathbf{51.81}$ & $\mathbf{45.79}$ & & $\mathbf{40.03}$ & $63.29$ \\
    $30$ & & $6$  & $62.19$ & $48.52$ & $43.53$ & & $39.9$ & $64.32$ \\
    \bottomrule
    \end{tabular}
    \caption{\textbf{Results of different numbers of Gaussian for video and text.}}
    \label{tab:mu}
    \end{minipage}
    \hspace{0.03\textwidth}
    \begin{minipage}{0.5\textwidth}
    \vspace{-10pt}
    \makeatletter\def\@captype{table}
    \centering
    \footnotesize
    \setlength{\tabcolsep}{5.2pt}
    \begin{tabular}{@{\hspace{0.4cm}} c@{\hspace{0.4cm}} c c@{\hspace{0.4cm}} c c c c c c}
    \toprule
    \multirow{3}{*}{\vspace{-0.2cm}$\lambda_{cta}$} & &\multirow{3}{*}{\vspace{-0.2cm}$\lambda_{vld}$} & \multicolumn{3}{c}{\textbf{MR}} & & \multicolumn{2}{c}{\textbf{HD}}
    \\
     \cmidrule{4-6} \cmidrule{8-9}
    & & & R1 & R1 & mAP & & \multirow{2}{*}{mAP} & \multirow{2}{*}{HIT@1} \\
    & & & @0.5 & @0.7 & Avg. & & & \\
    \midrule
    $0.3$& &$0.7$  & $63.61$ & $48.19$ & $43.68$ & & $39.88$ & $63.68$ \\
    $0.5$ & &$1.0$  & $64.84$ & $50.13$ & $44.69$ & & $40.02$ & $\mathbf{64.13}$ \\
    $\mathbf{0.5}$ & & $\mathbf{0.5}$ & $\mathbf{65.10}$ & $\mathbf{51.81}$ & $\mathbf{45.79}$ & & $\mathbf{40.03}$ & ${63.29}$ \\
    $1.5$& & $0.7$  & $64.13$ & $49.81$ & $43.97$ & & $39.96$ & $63.48$ \\
    \bottomrule
    \end{tabular}
    \caption{\textbf{Results of the different hyper-parameters in Multi-Aspect Contrastive Learning.}}
    \label{tab:lambda}
    \end{minipage}
    \begin{minipage}{0.4\textwidth}
    \makeatletter\def\@captype{table}
    \centering
    \footnotesize
    \setlength{\tabcolsep}{5pt}
    \vspace{2.5mm}
    \begin{tabular}{c c c c c c c c c}
    \toprule
    & \multirow{3}{*}{\vspace{-0.2cm}Iterations} & & \multicolumn{3}{c}{\textbf{MR}} & & \multicolumn{2}{c}{\textbf{HD}}
    \\
     \cmidrule{4-6} \cmidrule{8-9}
      & & & R1 & R1 & mAP & & \multirow{2}{*}{mAP} & \multirow{2}{*}{HIT@1} \\
      & & & @0.5 & @0.7 & Avg. & & & \\
    \midrule
     & $3$ & & $63.55$ & $48.52$ & $43.19$ & & $40.03$ & $64.26$ \\
    & $4$ & & $63.48$ & $50.0$ & $44.09$ & & $39.77$ & $\mathbf{64.71}$ \\
    & $\mathbf{5}$ & & $\mathbf{65.10}$ & $\mathbf{51.81} $ & $\mathbf{45.79}$ & & $\mathbf{40.03}$ & $63.29$ \\
    & $6$ & & $62.65$ & $47.94$ & $43.09$ & & $39.44$ & $62.77$ \\
    \bottomrule
    \end{tabular}
    \caption{\textbf{Results of different iterations of EM Attention.}}
    \vspace{5pt}
    \label{tab:iter}
    \end{minipage}
    \hspace{0.1\textwidth}
    \begin{minipage}{0.42\textwidth}
    \makeatletter\def\@captype{table}
    \centering
    \footnotesize
    \hspace{2.5mm}
    \setlength{\tabcolsep}{4.5pt}
    \begin{tabular}{c@{\hspace{0.4cm}} c c@{\hspace{0.4cm}} c c c c c c}
    \toprule
    & \multirow{3}{*}{\vspace{-0.2cm}Layer Num.} & & \multicolumn{3}{c}{\textbf{MR}} & & \multicolumn{2}{c}{\textbf{HD}}
    \\
     \cmidrule{4-6} \cmidrule{8-9}
      & & & R1 & R1 & mAP & & \multirow{2}{*}{mAP} & \multirow{2}{*}{HIT@1} \\
      & & & @0.5 & @0.7 & Avg. & & & \\
    \midrule
    & $2$ & & $61.74$ & $46.52$ & $41.18$ & & $39.81$ & $63..1$ \\
    & $\mathbf{3}$ & & $\mathbf{65.10} $ & $\mathbf{51.81} $ & $\mathbf{45.79} $ & & $\mathbf{40.03}$ & $\mathbf{63.29}$ \\
    & $4$ & & $63.68$ & $49.61$ & $43.8$ & & $39.63$ & $62.65$ \\
    & $5$ & & $63.55$ & $49.16$ & $43.73$ & & $39.56$ & $62.77$ \\
    \bottomrule
    \end{tabular}
    \caption{\textbf{Results of different numbers of encoder and decoder layers.}}
    \label{tab:layer}
    \end{minipage}

\end{minipage}
\vspace{-20pt}
\end{table*}

\section{Additional Details on Experiment Settings}
\subsection{Datasets}
\label{sup:datasets}
QVHighlight is the most popular publicized dataset which consists of over 10,000 videos with human-written free-form text descriptions for moment retrieval and highlight detection. 
Charades-STA and TACoS are both for moment retrieval where Charades-STA comprises 16,128 query-moment pairs for indoor activities and TACoS contains 127 annotated videos from cooking scenarios.
TVSum and YouTube Highlights cater for highlight detection. 
Each of which includes 10 domains with 5 videos and 6 domains with 433 videos respectively.

\subsection{Metrics}
\label{sup:metrics}
Recall@1 with IoU thresholds 0.5 and 0.7, mean average precision (mAP) with IoU thresholds 0.5 and 0.75 as well as average mAP over 0.5:0.05:0.95 are for MR, while mAP and HIT@1 are used for HD. HIT@1 is computed through the hit ratio of the clip with the highest score.
For Charades-STA and TACoS, we report the result of Recall@1 with IoU thresholds 0.5 and 0.7.
For YouTube Highlights and TVSum, we follow~\cite{umt} and adopt the metrics of mAP and Top-5 mAP.

\subsection{Feature Representations}
\label{sup:features}
The pre-extracted visual and text features from SlowFast~\cite{slowfast} and CLIP~\cite{clip} are used on all datasets.
Notably, on Charades-STA and YouTube Highlights, we additionally extract the features from official VGG~\cite{vgg} as well as GLOVE~\cite{glove} embeddings and commonly used I3D~\cite{i3d}, respectively for further comparisons. Besides, we leverage PANN~\cite{pann} model to encode audio features for experiments with extra audio modality learning.

\subsection{Training Details}
\label{sup:training details}
Elaborate parameter settings for each benchmark are summarized in \cref{tab:sup_training_details}. For more details,  
all experiments are implemented in PyTorch with one 24GB RTX3090.
The overall aggregation for DBIA is performed in $5$ iterations. In addition, the hidden dimension of transformer is $256$ for all experiments.

\section{Additional Experiments}

\subsection{Result on Charades-STA}
\label{sup:result_cha_vgg}
We also present comparisons on Charades-STA~\cite{cha} with existing methods in \cref{tab:cha_vgg}. As observed, our UVCOM achieve the new state-of-the-art performance under different settings, which further validates the rationality of our design in local perception enhancement for MR.

\subsection{Additional Ablation Studies}
We conduct additional analysis experiments on the val split of QVHighlights benchmark.
\begin{figure*}[ht]
    \centering
    \includegraphics[width=\linewidth]{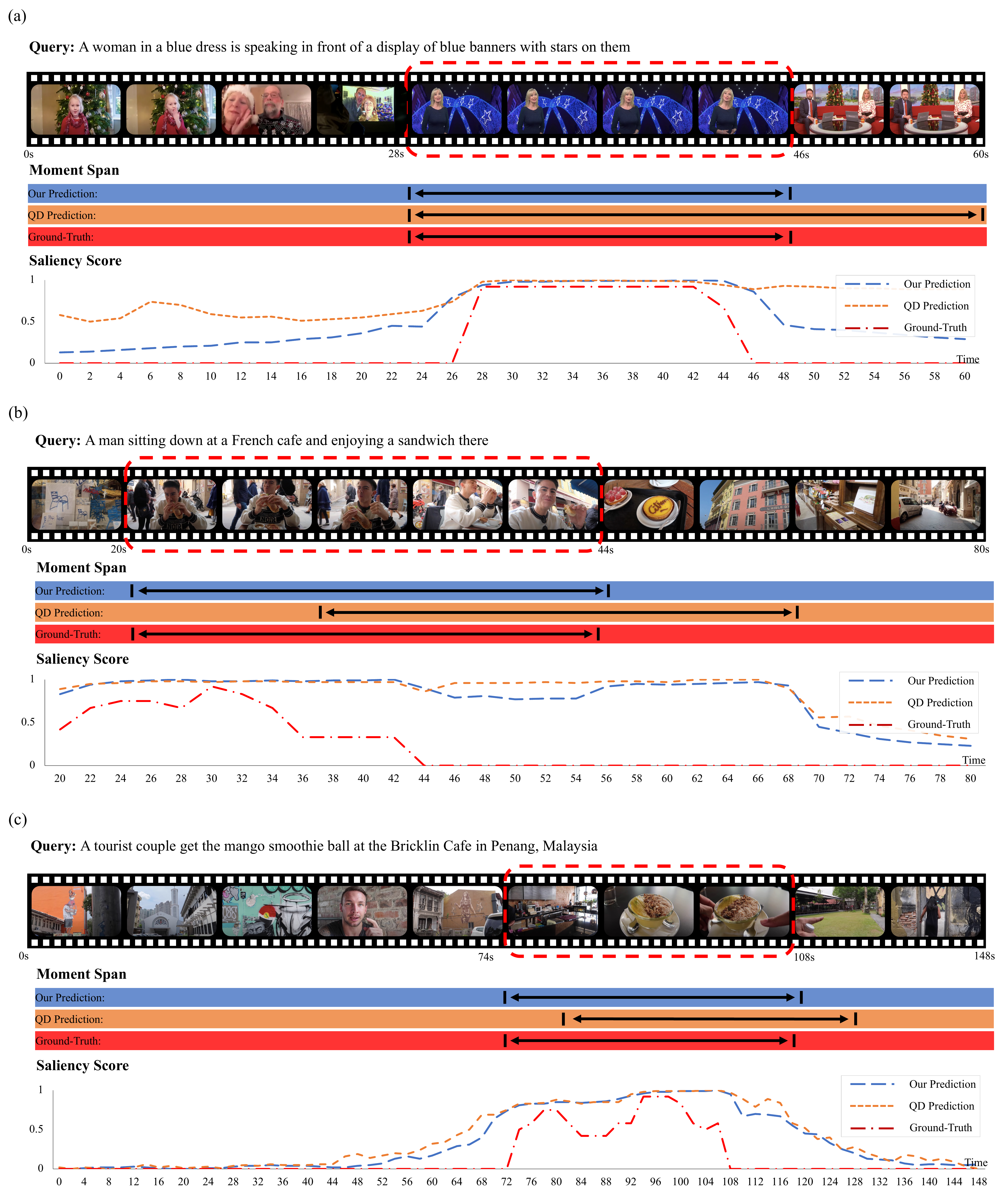}
    \caption{\textbf{Visualization comparison on MR and HD.} QD indicates previous state-of-the-art method QD-DETR~\cite{qddetr}}
    \label{fig:sup_case_show}
\end{figure*}

\begin{figure*}[ht]
    \centering
    \includegraphics[width=\linewidth]{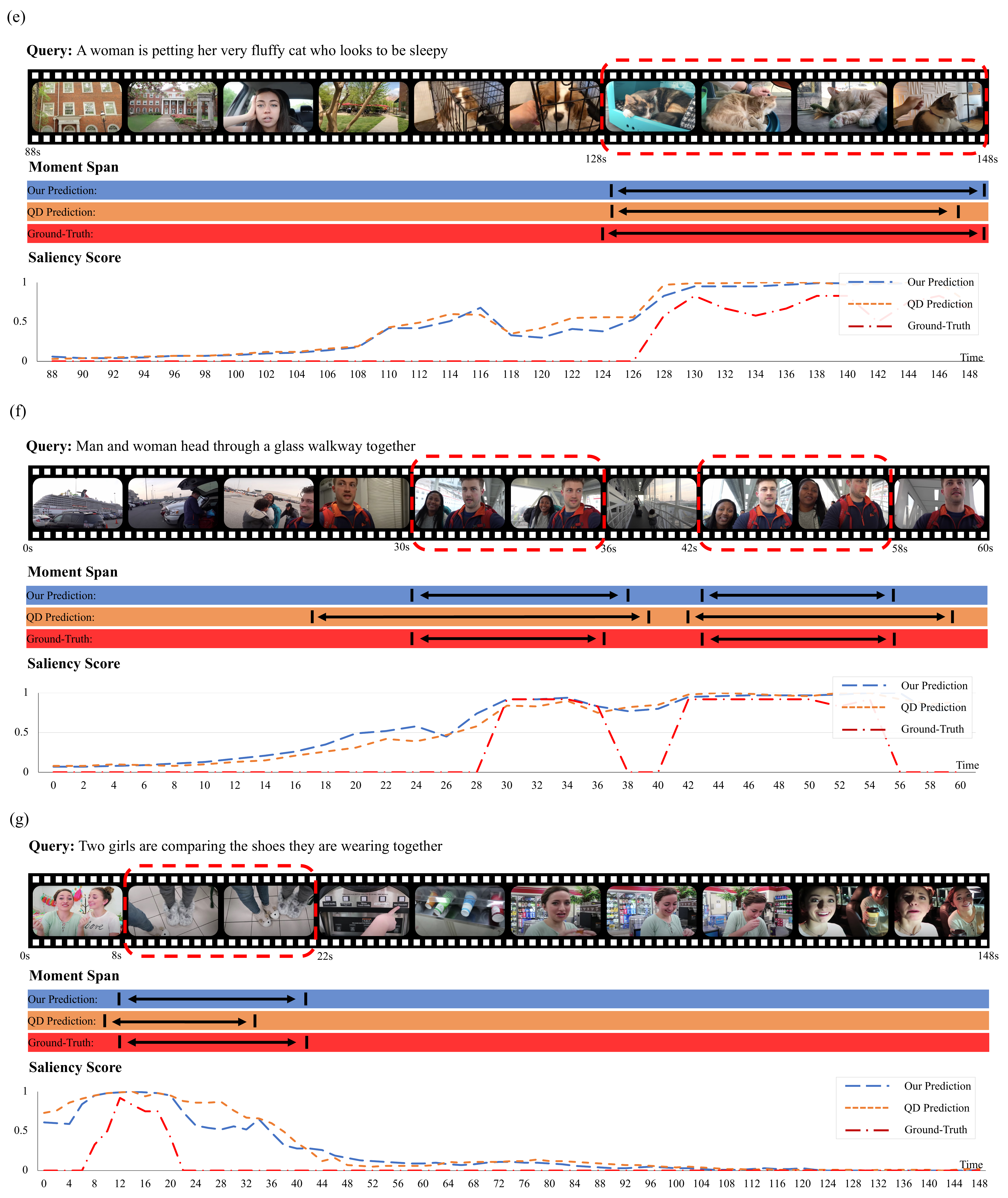}
    \caption{\textbf{Visualization comparison on MR and HD.} QD indicates previous state-of-the-art method QD-DETR~\cite{qddetr}}
    \label{fig:sup_case_show_2}
\end{figure*}
\paragraph{Number of Gaussian.}
The degree of the intra-modality aggregation is determined by the number of Gaussian. Specifically, we conduct ablation studies on two parameters $n_v$ and $n_t$ where each denotes the number of condense visual and linguistic outputs in \cref{tab:mu}. We find that the performance tends to boost as the number of Gaussians increases for which the smaller one may lead to inefficiency for content clustering in video or text. However, the large value of Guassians degrades the performance. We assume that the larger one introduces redundancy, which hinders the converge process. 
\begin{table*}[t]
\begin{minipage}[c]{\textwidth}
    \begin{minipage}{0.35\textwidth}
    \makeatletter\def\@captype{table}
    \centering
    \footnotesize
    \setlength{\tabcolsep}{2pt}
    \begin{tabular}{c | c c c c c c c c c c}
    \toprule
    Domain & VT & VU & GA & MS & PK & PR &FM & BK & BT & DS \\
    \midrule
    $\lambda_{cta}$ & 0.4 & 0.75 & 0.5 & 0.1 & 0.3 & 1.0 &0.25 &0.5 &1 &0.25 \\
    \midrule
    $\lambda_{vld}$ & 0.2 & 0.75 & 0.5 & 0.1 & 0.3 & 0 &0.25 &0.5 &0 &0.25 \\
    \bottomrule
    \end{tabular}
    \caption{\textbf{$\lambda_{cta}$ and $\lambda_{vld}$ for TVSum.}}
    \label{tab:sup_tvsum_loss}
    \end{minipage}
    \hspace{0.04\textwidth}
    \begin{minipage}{0.3
    \textwidth}
    \makeatletter\def\@captype{table}
    \centering
    \footnotesize
    \setlength{\tabcolsep}{2pt}
    \vspace{10pt}
    \begin{tabular}{c | c c c c c c}
    \toprule
    Domain & Dog & Gym. & Park. & Ska. & Ski. & Surf. \\
    \midrule
    $\lambda_{cta}$ & 0.5 & 0.8 & 0 & 0.5 & 0 & 0.5 \\
    \midrule
    $\lambda_{vld}$ & 1 & 0.8 & 1.5 & 1 & 1.5 & 1 \\
    \bottomrule
    \end{tabular}
    \caption{\textbf{$\lambda_{cta}$ and $\lambda_{vld}$ for YoutubeHL using SF+C feature.}}
    \label{tab:sup_youtubehl_sfc_loss}
    \end{minipage}
    \hspace{0.01\textwidth}
    \begin{minipage}{0.3\textwidth}
    \makeatletter\def\@captype{table}
    \centering
    \footnotesize
    \vspace{10pt}
    \setlength{\tabcolsep}{2pt}
    \begin{tabular}{c | c c c c c c}
    \toprule
    Domain & Dog & Gym. & Park. & Ska. & Ski. & Surf. \\
    \midrule
    $\lambda_{cta}$ & 0.5 & 0.8 & 1.0 & 0.5 & 1.0 & 0.5 \\
    \midrule
    $\lambda_{vld}$ & 0.5 & 0.8 & 1.5 & 1 & 1.5 & 0.5 \\
    \bottomrule
    \end{tabular}
    \caption{\textbf{$\lambda_{cta}$ and $\lambda_{vld}$ for YoutubeHL using I3D feature.}}
    \label{tab:sup_youtubehl_i3d_loss}
    \end{minipage}
\end{minipage}
\end{table*}
\paragraph{Multi-Aspect Contrastive Learning Coefficients.}
For multi-aspect contrastive learning, $\mathcal{L}_{cta}$, $\mathcal{L}_{vld}$ are coefficients for clip-text alignment and video-linguist discrimination loss respectively. We report the ablation results in \cref{tab:lambda}. As observed, the appropriate setting of coefficient helps decently solidify the local relation modeling and global knowledge integration, thereby facilitating the comprehensive understanding. 

\paragraph{Aggregation Iteration.} The aggregation iteration controls the quality of multi-grained feature generation, \textit{i.e.}, moment-level and phrase-level features. It can be seen in \cref{tab:iter} that the insufficient or excessive iterations both lead to the decreased performance due to the incomplete contextual information fusion or over exaggeration on similar contents. 

\paragraph{Layers.} To investigate the impact on the different encoder and decoder layers, we provide the performance variation in \cref{tab:layer}. The results depict that the increased layers bring significant improvement. However, the performance saturates when the number of layers reaches $5$. This can be attributed to the noises accumulation caused by redundant interaction.

\subsection{Visualizations.}
\cref{fig:sup_case_show} and~\cref{fig:sup_case_show_2} display additional qualitative comparisons between our UVCOM and a previous state-of-the-art method, which shows our consistent performance on Moment Retrieval and Highlight Detection.


\end{document}